# Digital Video Manipulation Detection Technique Based on Compression Algorithms

Edgar González Fernández, Ana Lucila Sandoval Orozco, and Luis Javier García Villalba, *Senior Member, IEEE*

*Abstract*—Digital images and videos play a very important role in everyday life. Nowadays, people have access the affordable mobile devices equipped with advanced integrated cameras and powerful image processing applications. Technological development facilitates not only the generation of multimedia content, but also the intentional modification of it, either with recreational or malicious purposes. This is where forensic techniques to detect manipulation of images and videos become essential. This paper proposes a forensic technique by analysing compression algorithms used by the H.264 coding. The presence of recompression uses information of macroblocks, a characteristic of the H.264-MPEG4 standard, and motion vectors. A Vector Support Machine is used to create the model that allows to accurately detect if a video has been recompressed.

## I. Introduction

FOR centuries, human beings have used images to shape the reality around them, or to modify it, depending on the message they want to convey. The invention of photography in the 19th century is a turning point in the use of images. According to Collier [1]: *"The excitement that greeted the invention of photography was the sense that man for the first time could see the world as it really is."*

However, this statement may not be completely accurate in today's digital age. The ease with which digital images and videos can be manipulated has increased dramatically in recent times, and several software and mobile applications to perform modifications, such as Adobe Photoshop, GIMP, Adobe Premiere, Snapchat, etc., are available to conventional users. These manipulations might be useful for improvement purposes, and they employ frequently tools based on artificial intelligence to achieve these improvements, such as object and facial detection and recognition, and augmented reality, among others.

In addition to this, due to the ubiquity of mobile devices and small cameras, it is possible to find a significant number of computer-related offences involving illegal possession, distribution or modification of multimedia content. The use of mobile devices for this purpose makes them an important source of evidence, which is why forensic analysis must be able to authenticate the content and examine whether it is original or has been manipulated.

The capabilities that modern technology offers to deceive human sight was explore researchers from the journal Cognitive-Research [2] in July 2017. They used a dataset of 40 scenes, 30 of which were subjected to five different types of manipulation, including physically plausible and implausible manipulations. These were shown to 707 participants, who were asked to distinguish whether a scene had been manipulated or not. The study found that only 60% of the people were able to detect the fake scenes, and even then, only 45% of them were able to tell exactly where the content alteration was.

All these arguments, are the basis of a strong motivation to develop tools to accurately detect modifications in images and videos. Methods for verifying the authenticity and integrity must be updated constantly, following new methods and trends of media modification and image processing.

This work is structured as follows: a brief introduction to the techniques of manipulation in digital videos is given in Section II. In Section III, a summary a of related works involving video manipulation detection is presented. Basic concepts of video compression, are explored in IV. The proposed detection algorithm and mathematical basic concepts required to develop this work are explained in Section V. Later, in Section VI, the experiments carried out and the results obtained are shown. Finally, the conclusions and proposals for future work in this field are presented in Section VII.

## II. Manipulation Techniques in Videos

In a general way, videos consists of a sequence of images called frames. The process of codification to achieve good levels of compression, depend on the techniques used to take advantage of the spatial and temporal redundancy that

Manuscript received December 26, 2020; revised February 21, 2021; accepted March 11, 2021. Date of publication December 30, 2021; date of current version March 9, 2022. This work was supported by the European Union's Horizon 2020 Research and Innovation Program under Agreement 101021801. The Associate Editor for this article was T.-H. Kim. *(Corresponding author: Luis Javier García Villalba.)*

Edgar González Fernández is with INFOTEC, Center for Research and Innovation in Information and Communication Technologies, Aguascalientes 20313, Mexico, and also with the Group of Analysis, Security and Systems (GASS), Department of Software Engineering and Artificial Intelligence (DISIA), Faculty of Computer Science and Engineering, Office 431, Universidad Complutense de Madrid (UCM), 28040 Madrid, Spain (e-mail: edgar.gonzalez@ucm.es).

Ana Lucila Sandoval Orozco and Luis Javier García Villalba are with the Group of Analysis, Security and Systems (GASS), Department of Software Engineering and Artificial Intelligence (DISIA), Faculty of Computer Science and Engineering, Office 431, Universidad Complutense de Madrid (UCM), 28040 Madrid, Spain (e-mail: asandoval@fdi.ucm.es; javiergv@fdi.ucm.es).

Digital Object Identifier 10.1109/TITS.2021.3132227

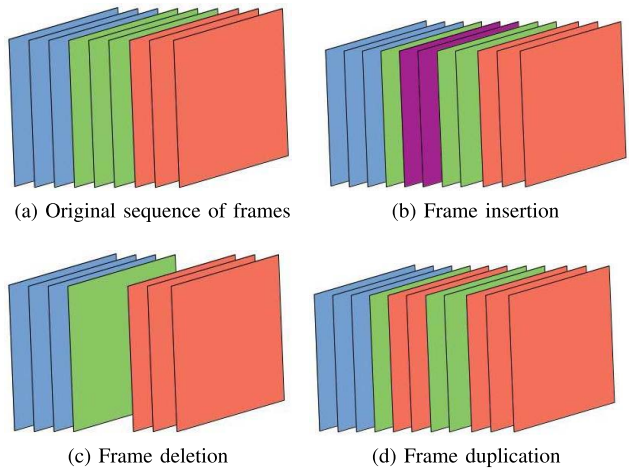

Fig. 1. Inter-frame manipulations.

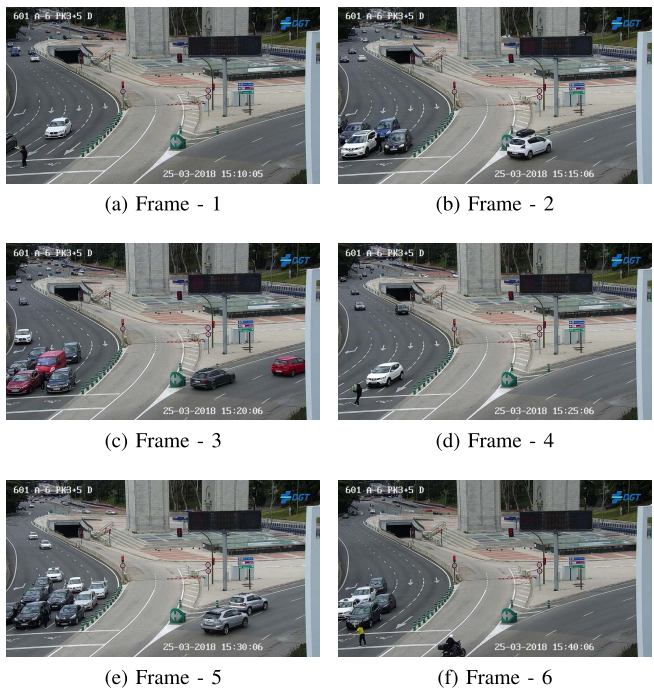

Fig. 2. Example of frames from a security camera.

contiguous frames have. A first classification of modifications on videos considers whether the modification alters the spatial or temporal information, by altering partially or totally the images extracted from individual frames.

### A. Inter-Frame Manipulations

Inter-frame manipulations consist in modifications of the original information of an image, by inserting, duplicating, exchanging or deleting complete frames that make up the video, as shown in Fig. 1.

One of the main purposes of this manipulation, is to remove an undesired event registered in the recorded scene. Additionally, it is also possible to include objects, or actions in the scene by adding external frames. If we take as an example the sequence of surveillance images from a traffic camera, such as those in Fig. 2, it is easy to make the white vehicle in Fig. 2d disappear from the scene by removing that frame.

Even though human eye is generally incapable of detecting differences between an original video and a video with inter-frame modifications, specially when modifying a small amount of frames, the manipulation processing operations leave a trace in the content information after compression algorithms take place.

### B. Intra-Frame Manipulations

Intra-frame manipulation focuses on the alteration of individual frames, and can be classified as follows:

- **Pixel-level manipulation**: This consists of treating the frame as a single image and applying manipulation techniques to images such as the views in the previous section, for example, copy-move, splicing, or in-painting.
- **Frame level manipulation**: In this case a modification affecting the whole frame is employed, such as resize or edge cropping. This can be used to hide video content that is located at the edges of the frame. For example, time and place marks.

Unlike the example shown with the inter-frame techniques, if the objective is to hide the presence of a vehicle from a surveillance camera (Fig. 2), an intra-frame techniques would target edges, either by removing them, or applying a copy-move modification, instead of removing the frames in which the car appears.

## III. PREVIOUS WORK

### A. Intra-Frame Manipulation Detection

In general, techniques applied to disclose manipulations in images, can be applied to individual frames to detect intra-frame modifications, such as copy-move, splicing, or in-painting, among others.

In-painting is a technique that has seen an increasing attention due to the improved results obtained by using novel machine-learning and artificial intelligence algorithms [3]–[6] and its use in specialised software applications. A technique to disclose in-painting, is given in [7]. A video is analysed by subtracting pixels that occupy the same spatial position in consecutive frames. In order to detect the duplicated content, authors correlate the 3D blocks of the frames. The presence of high correlation indicates the location of identical content in contiguous frames. Another technique for in-painting detection is provided in [8]. Individual frames are analysed, and a *zero-connectivity* feature is employed together with a fuzzy membership function, which relates blocks that have been created artificially, thus lacking of image artefacts present in authentic zones of the video, and introduced through the process of image formation.

In [9], authors proposed an approach for detecting and locating regionally-based in-painting in videos. The method detects irregularities in spatial-temporal coherence between consecutive frames. The video is first divided into sets of frames and then the coherence between each of these sets is calculated. In other words, sets with unnaturally high coherence or abnormally low coherence would be classified as manipulated frames.

Another good source of information when dealing with images, is the noise produced by the originating device,

as noted in [10]–[12]. In the former work, a multi-modal fusion protocol is applied to noise of frames, using a combination different correlation residue features, obtained after performing latent semantic, cross-modal factor, and canonical correlation analyses on the noise sub-blocks. The remaining works, use temporal cross-correlation applied to the noise residues discloses modifications by detecting unusual values of the correlation in the tampered areas. Aggressive lossy compression applied in compression techniques poses a big difficulty for noise estimation from video frames, which might make this algorithms suitable for surveillance videos.

In [13], authors observed that re-sampling caused by applying affine transformations to a video, introduces certain statistical correlations about the given content on individual frames. The correlations exposes modified zones, but requires of noise estimation as well. Another related work, but focused on images, can be found in [14], where an Expectation-Maximisation algorithm, together with Weighted Least Squares, allows to detect re sampling and estimate the kernel for the transformation.

### B. Inter-Frame Manipulation Detection

Devices introduce noise into each frame when recording a video. Since this noise follows a particular pattern in a sequence of consecutive frames, it is possible that, by inspecting these traces, changes between the frames are detected.

In [15], authors used the variance between the average noise of the frames and a particular one. Frames with higher variances would be marked as inserts. It was not proven to be effective on compressed video, and furthermore, it was tested on self-recorded video and is not sufficient to determine its applicability.

Authors of [16] proposed an adaptive motion algorithm that was able to detect and locate falsifications in interlaced and de-interlaced videos. They were based on the detection of correlation disturbances for interlaced and inter-frame motion disturbances for de-interlaced. This method, however, was ineffective for low-quality video.

Reference [17] uses the concept of the camera's Sensor Pattern Noise (SPN) to determine if all the frames in the video were recorded with the same device. The results obtained indicated that the algorithm was reliable for uncompressed video, but performance deteriorated for compressed video.

Another way to manipulate a video is by temporarily cutting it, interspersing frames from two different videos. When merging frames from different videos, it must be taken into account that it is necessary to synchronise their speeds (frame-rate).

The method suggested in [18] is based on the motion-compensated interpolation property because it leaves detectable traces in the frames. The authors were able to suggest a system that worked for uncompressed and slightly compressed video (e.g., H.264, or TV broadcast video) and achieved promising results, even when used on only a subset of frames. In addition, the system worked well in small spatial windows, allowing this detector to be used as a possible tool for detecting copy-paste counterfeit attacks. However, the number of interpolated frames observed had to be large enough for the system to successfully detect counterfeits.

In [19] the up-rate conversion of frames based on edge strength is detected. They use a certain threshold to distinguish the original zones from those that are up-converted, and based on this, they estimate the theoretical speed of the original frames. For a total of 300 test sequences, they achieved an average detection rate of 95%.

The authors in [20] developed a blind detection method based on frame-level analysis of a characteristic called 'mean texture variation' (MTV). Each generated ATV curve was processed in the video candidate as evidence of the upward velocity conversion. This technique could locate the position of the interpolation of the frames and help estimate their original speed.

### C. Recompression Detection

Recompression, or double compression, is an inevitable consequence of counterfeiting, since, after applying successfully a modification, the compression process must be applied again to save changes.

The first steps in this direction can be traced back to [21], where authors propose an algorithm based on the simple assumption that when an MPEG video is manipulated, two compressions took place: first, when the video was created, and second, when it was resaved after such alteration. They also exploited the fact that within a Group of Pictures (GOP), frames show a high correlation between them, so that adding or deleting a frame in a GOP increases the motion estimation error, which also results in periodic detectable peaks.

Later, another work by Su *et al.* [22], also focused on detecting frame-based tampering by detecting double compression in MPEG-2 videos. Instead of basing the frame aggregation/deletion detection process on temporal characteristics, the authors suggested using frequency characteristics. It was observed that when a video is recompressed after the frame is added/deleted, some high frequency components are lost in the recompressed frames due to the desynchronisation of the GOP and the non-linear quantification performed in the encoding process. These variations not only help to detect the forgery but also to locate it.

Another counterfeit detection technique based on double MPEG compression is the one proposed in [23], where abnormalities in the DCT coefficient patterns are treated as an indication of frame insertion/deletion. The authors extracted characteristics from the GOP, which are then used by a Support Vector Machine (SVM) to determine the original bit rate of the given double-compressed video, and it is observed that the detection performance of the technique was relatively lower for videos with lower bit rates, because a larger quantification scale requires a more robust quantification process, which the technique was not prepared to handle.

In the same year, a similar technique was pre-planned in [24], although with a novelty: its ability to detect transcoded videos, that is, videos that had been double compressed using two different compression standards. The authors also observed that after an MPEG-2 video was transformed into MPEG-4 video, the previous MPEG-2 compression traces, these generated new periodicities that were clearly observed in the histograms of the reconstructed Discrete Cosine Transform (DCT) coefficients. The authors presented the results in the form of receiver operating characteristic curves and stated that perfect results were obtained in case of low bit rates. These curves also showed that as the target output bit rate increased, the detection performance decreased. It also assumed that transcoding always suggested manipulation.

On the other hand, in [25] it is proposed to detect double encoding even if the main frame set had been removed. This method has the additional advantage of being able to effectively locate the counterfeit, and is also suitable for H.264 encoded videos, as opposed to [24] that worked only for Moving Picture Experts Group (MPEG) videos. The modified methodology was also able to estimate the number of deleted frames.

Reference [26] uses the Markov statistics to detect double compression. They rely on the fact that double quantisation with different parameters will inevitably introduce rounding errors, leaving artefacts detectable. The random Markov process could capture such artefacts for detection. In [27] author's approach is based on the extraction statistical characteristics of the macroblocks of P-frames. They propose to detect double MPEG compression with the same QS. Feature extraction occurs during repeated compression of the video at the same QS factor.

The problem of estimating the exact number of compressions is addressed in [28]. In this work, a combination of multiple SVM is applied on feature vectors, created by considering statistical information about the most significant digit of the quantised transformed coefficients, and justifying their theoretical assumptions by what the so-known Benford's Law and its applications to image processing and digital forensics [29], [30].

Aside from analysis made directly to the content of the video, metadata might provide useful information of post-processing [31]–[34], for example, by looking for information left by editing software, changes in the structure of files obtained from the source device, inconsistencies in metadata information (frame rate, size, duration, etc.), or differences in thumbnails against full size images. The main drawback of this approach is that experienced users might edit this information to conceal the modifications made. Nevertheless, techniques based on metadata analysis is useful in many cases.

A warning is made in [35], where authors state that multiple compressions were an under-explored topic and that it is risky to make assumptions regarding the authenticity of digital content simply on the basis of the presence of double compression. Their claim was supported by the simple fact that digital content available on the Internet generally undergoes more than one compression, even if video is only transmitted, uploaded, downloaded or watched. Many of these works have been carried out for MPEG-2 standard, and for this reason, the work presents here aims to address the very common H.264 compression. For this reason, some techniques address double compression by analysing DCT coefficients, assuming images are compressed using JPEG-like algorithms. However, H.264 introduces new compression techniques [36]. The main concepts, useful for the detection technique introduced in Section V will be explained next.

## IV. THE H.264/MPEG VIDEO COMPRESSION

Given that H.264, also known as MPEG-4 Part 10, is widely used by several devices, internet streaming sources, and also TV transmissions, the analysis focuses on modifications made on videos encoded with this standard. Let us remind some of the main and concepts involved in the H.264 compression standard.

### A. Frame Types

A frame can be considered as an individual image which contains a small amount of the temporal information contained in a full length video. The whole set of frames is presented in sequence, at a speed such that the transition from a frame to the next is not noticed by the human eye, and gives the feeling of a continuous and smooth movement. The information that two temporally close frames have, introduces a lot of redundancy. This characteristic is used by compression techniques to reduce file sizes. In the case of H.264, frames are pre-processed to take advantage of this redundancy by classifying them as follows, according to how their information will be compressed and encoded:

- **I-Frames**: they are encoded using only information found within the same frame. Spatial redundancy can be used in order to achieve better compression results, and common techniques partition the image in blocks of size $16 \times 16$, or even at size $4 \times 4$ to achieve better quality.
- **P-Frames**: in addition to spatial redundancy, P-Frames also take advantage of the high temporal redundancy found between close frames. To achieve good levels of compression, motion estimation and compensation techniques are employed by partitioning the image in blocks. These type of frames, refer only to the information found in past frames.
- **B-Frames**: similar to P-Frames, but in this case, temporal redundancy from both, past and future frames is used.

Motion estimation techniques used by implementations of the H.264 standard require of partitioning a frame in blocks, which are individually processed (see Fig. 3). These blocks, known as *macroblocks* are also classified according to the procedure specified used to perform codification:

- **I-MB**: For I-Macroblocks (intra-prediction), only macroblocks within the same frame that have been reconstructed are used. According to the ITU H.264 recommendation, I-MBs can be processed either considering the original $16 \times 16$ size, or by using a $4 \times 4$ partition. This type of macroblock can be found in I, P or B-frames.
- **P-MB**: Predicted macroblocks use blocks from other frames to be estimated and encoded. In this case, a motion vector is associated to the macroblock, which identifies the location of the referenced block. Also, a motion compensation is computed, this is, the error of the estimation. These macroblocks can be found in P and B frames.
- **S-MB**: Skipped macroblocks, as the name suggests, are those for which no information is transferred to the

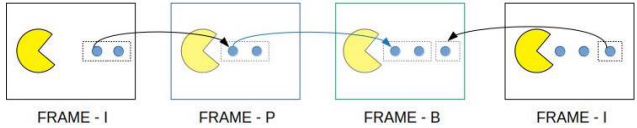

Fig. 3. Prediction sequence of frames according to type.

decoder. When decoding takes place, a macroblock must be selected to supply the information missing. This macroblocks can be found in P or B-frames.

P and B macroblocks can be further partitioned in smaller subblocks, of size $16 \times 8$, $8 \times 16$ or $8 \times 8$ to improve quality. However, during the motion compensation step, a vector for each subblock is required.

This is a rough classification, and it will be further refined in Section V-A. Let us continue with a shallow introduction of the prediction mode.

### B. Intra-Prediction

Estimation of blocks in intra-prediction mode is carried out based on previously encoded and reconstructed blocks (within the same frame). For the luminance sample, the whole $16 \times 16$ block can be processed, or it can be further partitioned in $4 \times 4$ subblocks. In the case of $16 \times 16$ prediction, 4 modes can be used, all of them using the information found in the left and upper blocks, and using a plane extrapolation. for the $4 \times 4$ partition, 9 different modes can be selected, using the information of left and top pixels as well. In both cases, selection is decided according to the mode that better approximates the original data.

Chromatic samples are always processed as $16 \times 16$ blocks using one of the 4 modes explained for the luminance case.

### C. Inter-Prediction

In the case of inter-prediction mode, blocks are predicted using information from blocks contained in previously reconstructed frames. According to the smoothness of the region, macroblocks can be further partitioned in $16\times 8$, $8\times 16$, or $8\times 8$ subblocks. These subblocks can also be further partitioned in $8 \times 4$, $4 \times 8$, or $4 \times 4$ subblocks, to achieve a more detailed result. The election of the subblock size involves a tradeoff between image fidelity, and data economy and processing time: each subblock requires of a motion vector, which in turn, must be selected from a spatial neighbourhood of a different frame.

Additionally, the motion compensation must be stored as well. In this case, smooth blocks give low energy residuals. Thus, big partitions will be used for large homogeneous areas, while areas with more texture will need smaller partitions.

## V. Multiple Compression Detection Technique

The detection technique is based on the analysis of the differences found in the macroblock types and motion vectors in a video and a recompressed version. For this purpose, the FFMPEG tool is used. We start this section by presenting an overview of the methods and information that can be extracted from a video with FFMPEG.

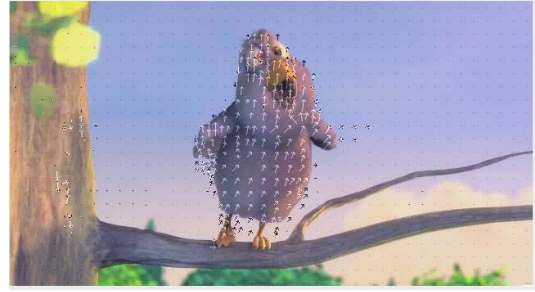

Fig. 4. Motion vectors extracted from a MPEG4 compressed video frame.

### A. The FFMPEG Project

FFMPEG is a free software multimedia platform capable of decoding, encoding, transcoding, transmitting, filtering and playing most audio and video formats. It is developed on GNU/Linux but also compiles and runs on most operating systems, development environments, architectures and configurations. It has a GNU LGPL license, which guarantees a certain freedom when sharing and modifying the software, ensuring that the software is free for all its users [37]. It is possible to use FFMPEG to analyse the macroblocks and motion vectors of any MP4 video file. An example of a frame with the analysed motion vectors printed as arrows can be seen in Fig. 4.

*1) Macroblock Types:* As mentioned in Section IV-A, frames and macroblocks are classified in types according to the way they are processed, compressed, and encoded. To obtain the macroblock types, it is possible to use the -debug option, together with the mb_type flag. This command outputs an matrix of macroblock types, with the most usual types being encoded as follows:

- i: $4 \times 4$ intra-prediction with 9 different modes of interpolation,
- D: skipped blocks (B slice),
- >: reference to previous frame (P or B slices),
- <: reference to future frame (B slices),
- X: reference to past and future (B slices).

Additionally, partitions are encoded in the debug output as follows:

- |: the macroblock is partitioned into two subblocks of size $16 \times 8$,
- -: the macroblock is partitioned into two subblocks of size $8 \times 16$,
- +: the macroblock is partitioned into four subblocks of size $8 \times 8$.

Once the types and partitions are obtained with the ffmpeg method, differences in block types, and motion vectors of each recompression are counted to compute the feature for the SVM. As more compressions are performed, fewer changes are found. To obtain the information of motion vectors, the tool MPEG-flow[1] has been used.

---

[1] https://devhub.io/repos/vadimkantorov-mpegflow

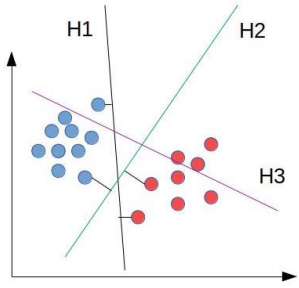

Fig. 5. Main elements of SVM.

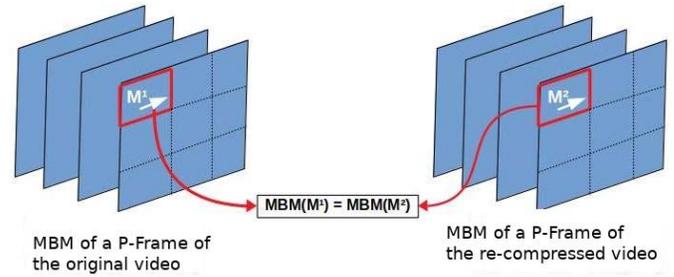

Fig. 6. Stable macroblock.

## B. Support Vector Machine

SVMs are supervised machine learning models, very useful for classification, pattern recognition, and regression analysis. Classification is performed using a sample of vectors from a vector space for the training phase. The objective of the SVM is to find the best hyper-plane that divides the data of all the training samples into two or more well-differentiated classes, that is, to determine the hyper-plane with the maximum distance from the point of each class that is closest to it (Fig. 5). The new data was grouped to the cluster to which the distance is less.

Nevertheless, two problems arise when employing SVM:
- Spaces studied have generally more than two dimensions and do not have a linear representation. This problem is solved with the representation by kernel functions, which project the information to a space of multidimensional characteristics by means of a non-linear mapping.
- Appropriate kernel parameters selection. To apply non-linear techniques, some parameters depending on the kernel function must be estimated (for our case, the Radial Base Function (RBF), with parameters $C$ and $\gamma$). To find the best test and training classification parameters, the parameter optimisation method is used.

## C. Multiple Compression Detection

This algorithm will be used for the forensic purpose of determining whether a video has undergone more than one compression, which is the first evidence that the video may have been manipulated in any way. The detection of recompressions is based on the study of the statistical characteristics of the Macroblock Mode (MBM).

The MBM is a feature that consists of the type of macroblock and motion vector. To extract this feature, a video is recompressed repeatedly on the same quality scale and then calculate the number of different MBM between two sequential compressions. Finally, these extracted statistics are used by the SVM to determine whether the video is original or whether it has been recompressed.

This method is inspired by the convergence of the Joint Photographic Experts Group (JPEG) coefficients computed when an image is compressed multiple times [38].

For a macroblock $M$, the MBM is composed of the two properties as follows:

$$\mathrm{MBM}(M) = (M_{type}, M_{mv})$$

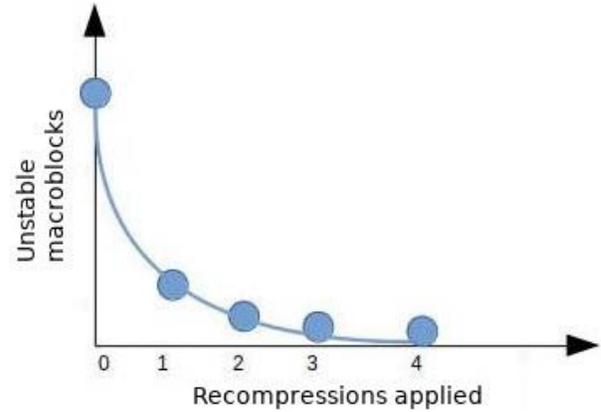

Fig. 7. Decrease in number of unstable macroblocks after successive recompressions.

where $M_{type}$ is the macroblock type and $M_{mv}$ is the motion vector. Two macroblocks are considered to have the same MBM if and only if $M_{type}$ and $M_{mv}$ are identical. Skipped and I-macroblocks do not require of a motion vector, thus, in these cases, comparison of $M_{mv}$ is not carried out.

For a sequence of compressions on a video, a macroblock in a given temporal and spatial position is *stable* if it has the same MBM in the $i$-th and the $(i+1)$-th compressions. Otherwise it is considered unstable. See Fig. 6.

The first part of the detection technique consists in extracting a single feature value from two H.264 videos. This feature registers the average number of unstable macroblocks detected in the frames of the whole video. To extend the analysis to $n$ compressions, a vector of $n$ entries are computed:

$$\mathbf{v}_f = (v_0, \ldots, v_{n-1bvfv})$$

where $v_i$ is the feature extracted from videos of the $i$-th and $(i+1)$-th recompressions. For $i = 0$, the original and first re-compression are considered. The obtained vector is the input for a previously trained SVM, which is used to estimate the number of re-compressions of the inspected video. For simplicity, the approach employed, extracts the feature by using only unstable macroblocks from P-Frames.

Even though the process can be applied for a specific number of re-compressions, accuracy drops as the number of re-compressions grows, as can be seen in Fig. 7. After a second compression, the average number of stable macroblocks drops slowly, making difficult to accurately determine the number of compressions.

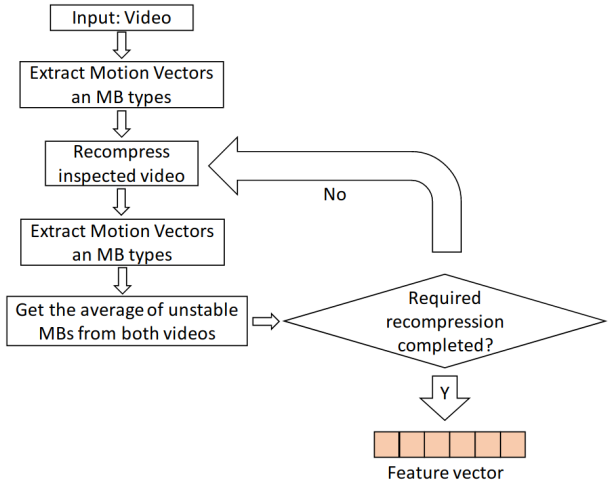

Fig. 8. Feature extraction process.

TABLE I
DATASET USED FOR EVALUATION

| Resolution | Format | Num. training videos | Num. prediction videos |
|---|---|---|---|
| 720×480 | MP4 | 20 | 20 |
| 720×1280 | MP4 | 40 | 35 |
| 1920×1080 | MP4 | 40 | 35 |
| 4K | MP4 | 20 | 17 |

TABLE II
COMPUTATIONAL RESOURCES

| Resources | Specifications |
|---|---|
| OS | Ubuntu 18.04 (64 bits) |
| Memory | 4 GB |
| Processor | Intel® Core™ 2 Quad CPU Q8200 @ 2.33GHz x 4 |
| Graphic card | NV96 |
| HD | 100 GB |

TABLE III
ACCURACY OF RE-COMPRESSED VIDEOS DETECTION

| Data types | Resolution | | | | |
|---|---|---|---|---|---|
| | 720×480 | 720×1280 | 1920×1080 | 4K | MIX |
| Scaled | 90% | 94,28% | 97,14% | 100% | 91,59% |
| Non-scaled | 90% | 100% | 100% | 100% | 92,52% |

To formalise the definition of the feature values $v_i$, let us define with $N$ the total number of P-frames in the video, and denote with $M_i(k, x, y)$ the macroblock found in the $i$-th re-compression, the $k$-th frame, and whose top-left corner starts at $(x, y)$. This means that $x, y$ are multiples of 16, the size of the macroblock.

Then computation of $v_i$ is given by Eq. (1).

$$v_i = \frac{1}{N} \sum_{k,x,y} I(M_i(k, x, y), M_{i+1}(k, x, y)) \quad (1)$$

where $I$ is the function defined by:

$$I(M_i(\mathbf{x}), M_j(\mathbf{x})) = \begin{cases} 0 & \text{if } \text{MBM}(M_i(\mathbf{x})) = \text{MBM}(M_j(\mathbf{x})) \\ 1 & \text{otherwise} \end{cases}$$

with $\mathbf{x} = (k, x, y)$. A dataset with multiple re-compressions of test videos is employed to create the training set using the feature extraction already explained.

The proposed detection algorithm can be better understood with the diagram shown in Fig.8.

## VI. EXPERIMENTS AND RESULTS

Throughout this section, all the experiments performed to evaluate the effectiveness of the training-based manipulation detection algorithm are shown. They are intended to check the variation in accuracy when applying the algorithm on different resolutions. The ability to detect whether the video is original, has had double compression or triple compression is studied.

To evaluate the proposed algorithm, a dataset containing digital videos from different models of mobile devices with different resolution sizes was used. The employed dataset has been generated using personal devices which allowed us to test the algorithm for different resolutions, and thus it is not publicly available by the time of writing. MP4 format videos have been selected with the most common resolutions: 720 × 480, 720 × 1280, 1920 × 1080 and 3840 × 2160 (4K). Most of the selected videos have been used for the training of the SVM in order to have a wider knowledge base. The rest have been used for prediction. Table I shows a summary of the characteristics of the dataset used for training and prediction.

The characteristics of the computer hardware on which the experiments have been performed are presented in Table II. This is an important factor to take into account since the execution times of the different tests may vary according to the available computer resources.

First, the knowledge base for training the vector support machine was created. The videos selected as training datasets for each resolution are used as input for the recompression detection algorithm. Once the machine is trained with the feature vectors generated by the algorithm, the prediction can be started. This test consists of extracting the characteristics of the videos to be tested so that the SVM machine, once trained, classifies them according to their recompressions.

In the first group of experiments, the aim is to detect whether a video is original or has had at least one recompression. In this case, the algorithm extracts only two characteristics from the videos of the dataset. Therefore, the vector support machine is trained with original and recompressed videos, the resulting model consists of two classes to discern whether the video is original or not. An experiment was run for the videos with the different resolutions of the dataset and additionally an experiment was done mixing all the resolutions to evaluate the tolerance of the algorithm to the size of the video. For each of these experiments the characteristic vectors were taken scaled (normalised) and unscaled. The results obtained for each of the experiments with different resolutions are presented in Table III.

As shown in the Table III, the proposed detection algorithm has a hit a detection rate above 90% even when the resolution is low (720 × 480). At all resolutions the results are superior when the feature vector is not scaled, even in the case where the system is trained with videos of different resolutions.

Tables IV and V show the resulting confusion matrices for each of the analysed resolutions with both, scaled and unscaled feature vectors respectively.

TABLE IV

CONFUSION MATRIX WITH SCALED DATA

| Resolution | Classes | Confusion Matrix | | Total Videos | Success Rate |
|---|---|---|---|---|---|
| | | Original | Double Comp. | | |
| 720×480 | Original | **10** | 0 | 10 | 90% |
| | Double Comp. | 2 | **8** | 10 | |
| 720×1280 | Original | **20** | 0 | 20 | 94,28% |
| | Double Comp. | 2 | **13** | 15 | |
| 1920×1080 | Original | **19** | 1 | 20 | 97,14% |
| | Double Comp. | 0 | **15** | 15 | |
| 4K | Original | **9** | 0 | 9 | 100% |
| | Double Comp. | 0 | **8** | 8 | |
| MIX | Original | **44** | 9 | 53 | 91,59% |
| | Double Comp. | 0 | **54** | 54 | |

TABLE V

CONFUSION MATRIX WITH UNSCALED DATA

| Resolution | Classes | Confusion Matrix | | Total Videos | Success Rate |
|---|---|---|---|---|---|
| | | Original | Double Comp. | | |
| 720×480 | Original | **9** | 1 | 10 | 90% |
| | Double Comp. | 1 | **9** | 10 | |
| 720×1280 | Original | **20** | 0 | 20 | 100% |
| | Double Comp. | 2 | **0** | 15 | |
| 1920×1080 | Original | **20** | 1 | 20 | 100% |
| | Double Comp. | 0 | **15** | 15 | |
| 4K | Original | **9** | 0 | 9 | 100% |
| | Double Comp. | 0 | **8** | 8 | |
| MIX | Original | **46** | 7 | 53 | 92,52% |
| | Double Comp. | 1 | **53** | 54 | |

TABLE VI

SUCCESS RATE FOR MULTIPLE RECOMPRESSION DETECTION

| Data types | Resolution | | | | |
|---|---|---|---|---|---|
| | 720×480 | 720×1280 | 1920×1080 | 4K | MIX |
| Scaled | 83,33% | 70% | 68% | 80% | 49,67% |
| Non-scaled | 66,67% | 88% | 70% | 60% | 63,23% |

In the second group of experiments one more feature is extracted than in the previous experiments in order to determine whether the analyzed videos have been recompressed more than once, and if so, to know if they have been recompressed one or two additional times. As in the previous case, for the experiments the characteristic vectors have been taken scaled and unscaled. The results obtained for each of the experiments with different resolutions are presented in Table VI.

As shown in the Table VI, the proposed detection algorithm has the best hit rate (88%) when the video has a resolution of 720 × 1280 and the data is unscaled. Tables VII and VIII show the resulting confounding matrices for each of the analyzed resolutions with the scaled and unscaled feature vectors, respectively.

Table IX shows the algorithm execution time for each of the resolutions.

Table X shows a comparison between the proposed method and related works found in the literature, and precision of each technique is detailed as reported in the original source. It is worth mentioning that results are applied on different datasets, which have less videos than the set considered in this work. Most of the state-of-art techniques are applied only to detect if a recompression has been carried out, thus, in the case of the results obtained in [28] and our proposal,

TABLE VII

CONFUSION MATRIX BY RESOLUTION WITH THREE CLASSES WITH SCALED DATA

| Resolution | Classes | Confusion Matrix | | | Total Vídeos | Success Rate |
|---|---|---|---|---|---|---|
| | | Original | Double | Triple | | |
| 720×480 | Original | **10** | 0 | 0 | 10 | 83,33% |
| | Double | 0 | **5** | 5 | 10 | |
| | Triple | 0 | 0 | **10** | 10 | |
| 720×1280 | Original | **20** | 0 | 0 | 20 | 70% |
| | Double | 6 | **9** | 0 | 15 | |
| | Triple | 1 | 8 | **6** | 15 | |
| 1920×1080 | Original | **20** | 0 | 0 | 20 | 68% |
| | Double | 0 | **0** | 15 | 15 | |
| | Triple | 0 | 1 | **14** | 15 | |
| 4K | Original | **9** | 0 | 0 | 9 | 80% |
| | Double | 0 | **6** | 2 | 8 | |
| | Triple | 0 | 3 | **5** | 8 | |
| MIX | Original | **25** | 2 | 20 | 47 | 49,67% |
| | Double | 0 | **0** | 54 | 54 | |
| | Triple | 1 | 1 | **52** | 54 | |

TABLE VIII

RESOLUTION CONFUSION MATRIX WITH THREE CLASSES WITH UNSCALED DATA

| Resolution | Classes | Confusion Matrix | | | Total Vídeos | Success Rate |
|---|---|---|---|---|---|---|
| | | Original | Double | Triple | | |
| 720×480 | Original | **10** | 0 | 0 | 10 | 66,67% |
| | Double | 1 | **5** | 4 | 10 | |
| | Triple | 0 | 5 | **5** | 10 | |
| 720×1280 | Original | **19** | 1 | 0 | 20 | 88% |
| | Double | 0 | **12** | 3 | 15 | |
| | Triple | 0 | 2 | **13** | 15 | |
| 1920×1080 | Original | **20** | 0 | 0 | 20 | 70% |
| | Double | 0 | **8** | 7 | 15 | |
| | Triple | 0 | 8 | **7** | 15 | |
| 4K | Original | **7** | 2 | 0 | 9 | 60% |
| | Double | 0 | **0** | 8 | 8 | |
| | Triple | 0 | 0 | **8** | 8 | |
| MIX | Original | **44** | 3 | 0 | 47 | 63,23% |
| | Double | 6 | **29** | 19 | 54 | |
| | Triple | 7 | 22 | **25** | 54 | |

TABLE IX

PERFORMANCE OF THE PROPOSED ALGORITHM

| Detected compression | 720×480 | 720×1280 | 1920×1080 | 4K |
|---|---|---|---|---|
| Double | 00:00:07.03s | 00:00:24.49s | 00:01:16.32s | 00:05:44.13s |
| Triple | 00:00:16.23s | 00:00:55.11s | 00:02:37.32s | 00:11:41.02s |

TABLE X

COMPARISON WITH STATE-OF-THE-ART TECHNIQUES

| References | Utilised Characteristics | Dataset | Precision | |
|---|---|---|---|---|
| | | | Double | Triple |
| [23] | First digit distribution of Alternating Current | 12 video set from [39] | 97,92% | – |
| [26] | Markov features and first digit distribution | Derf's collection [39] | >90% | – |
| [27] | Macroblock types of P-frames | Derf's collection [39] | 94,12% | – |
| [28] | Benford's law on quantised coefficients | Generated set of 12 videos | 100% | 77,78% |
| Proposed method | Macroblock types and motion vectors | Generated set of 227 videos | 95,27% | 75,33% |

the column for "Double Compression" shows the accuracy to detect if at least one recompression has been carried out. Observe that [28] outperforms our work, but we have used videos in several resolutions and sizes, which is not the case in the aforementioned work.

Other works mentioned in Section III have not been appended to Table X since no data of the accuracy of such methods has been explained in the inspected documents.

## VII. CONCLUSION

Digital image and video content has information that goes beyond the visual. This information is of great forensic value, since its correct exploitation can guarantee the authenticity and integrity of the content. Because of this, digital images and videos are an exceptional source of evidence when it comes to resolving legal proceedings. The development and continuous improvement of new technologies means that conventional users are capable of altering image and video content with professional results, imperceptible to the human eye. This is in addition to the fact that the detection of manipulations is a complex task and also requires continuous improvement to adapt to such a scenario, so it is essential to develop forensic tools capable of detecting these manipulations, which are increasingly professional and common.

The line of research that has been followed in this work begins with a study of existing techniques for detecting manipulation of digital images and videos, devoting more effort to techniques for detecting splices in images and double compression detection in videos.

A manipulation detection technique based on the H.264/MPEG4 video standard for the detection of recompressions in MP4 videos has been designed and implemented. This technique compares the motion vectors of the macroblocks of two sequential compressions of the same video, and then makes use of a SVM to classify the video.

A dataset has been created to evaluate the proposed recompression detection technique and a public dataset has been used to compare the results with other related research. The evaluation consisted of two experiments divided into groups according to the resolution of each video:

- Detection of original or double-compressed video, the algorithm has achieved maximum accuracy with data scaled to 100 percent for 4K resolution video, for all other resolutions not less than 90 percent.
- Detection of original video, double compression, or triple compression where the accuracy decreases slightly with respect to the detection of original or double compression, has an average accuracy of the order of 80%. However, the best result is still a high resolution.

The experiments have been carried out for both non-scaled and scaled data, obtaining very similar results between them. Therefore it is not relevant to scale them.

Tests have also been conducted by mixing all resolutions, obtaining less accurate results than in those tests where resolutions have been separated. The performance of the algorithm is directly proportional to the resolution of the video to be processed and the amount of recompressions to be detected.

Based on the results obtained in this research, the future lines of investigation proposed in this work are the following:

- Extend the recompression detection algorithm for use with video codecs other than H264.
- Use deep learning techniques and increase the number of extracted features to improve the accuracy of recompression detection.
- Optimise the recompression detection algorithm to reduce the processing time for high resolution video.

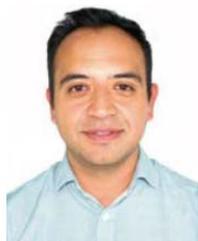

**Edgar González Fernández** received the M.S. degree in mathematics from the Mathematics Department, CINVESTAV, and the Ph.D. degree in computer science from the Computer Science Department, CINVESTAV, in 2020. He is currently a member of the Group of Analysis, Security and Systems (GASS), Universidad Complutense de Madrid (UCM). Since February 2020, he has been with INFOTEC, acting as an Assistant Manager of teaching, and since May 2020 as a Coordinator of the master's degree in data science. He is also a member of the National System of Researchers of the Mexican CONACyT. His areas of interest are cryptography, information security, data science, and image processing.

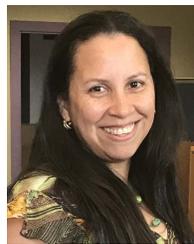

**Ana Lucila Sandoval Orozco** was born in Chivolo, Magdalena, Colombia, in 1976. She received the degree in computer science engineering from the Universidad Autónoma del Caribe, Colombia, in 2001, and the M.Sc. degree in research in computer science and the Ph.D. degree in computer science from the Universidad Complutense de Madrid, Spain, in 2009 and 2014, respectively. She completed a specialization course in computer networks at the Universidad del Norte, Colombia, in 2006. She is currently a Post-Doctoral Researcher and a member of the Group of Analysis, Security and Systems (GASS), Universidad Complutense de Madrid. She is also a Visiting Professor with the Network Engineering Laboratory (LabRedes), Electrical Engineering Department, University of Brasília, Brazil. She has been an advisor of seven Ph.D. theses in the last five years. Her professional experience includes her participation in both national and international research projects and both public and private financing. Regarding her scientific activity, she has published numerous publications in prestigious international conferences. Her main research interest is computer networks and their applications. She is also an Editor or a Guest Editor of international journals, such as *Applied Sciences* (MDPI), *Entropy* (MDPI), IEEE ACCESS, and the *International Journal of Distributed Sensor Networks*.

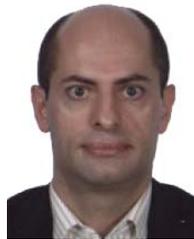

**Luis Javier García Villalba** (Senior Member, IEEE) received the degree in telecommunication engineering from the Universidad de Málaga, Spain, in 1993, and the Ph.D. degree in computer science from the Universidad Politécnica de Madrid, Spain, in 1999. He was a Visiting Scholar with the Research Group Computer Security and Industrial Cryptography (COSIC), Department of Electrical Engineering, Faculty of Engineering, Katholieke Universiteit Leuven, Belgium, in 2000, and a Visiting Scientist with the IBM Research Division, IBM Almaden Research Center, San Jose, CA, USA, in 2001 and 2002. He is currently an Associate Professor with the Department of Software Engineering and Artificial Intelligence, Universidad Complutense de Madrid (UCM). He is also the Founder and the Head of the Complutense Research Group, Group of Analysis, Security and Systems (GASS), Faculty of Computer Science and Engineering, UCM, and it is composed by a group of researchers and professors from different faculties (economics and business, mathematics, statistical studies, computer science and engineering, and the last one being the core faculty). GASS is clearly inter-departmental oriented, as the group members belong to different departments (applied mathematics, financial economics and accounting, information systems and computing, software engineering and artificial intelligence, and statistics and operations research III). As an interdisciplinary research team of UCM, GASS integrates mathematical solutions into different environments and applies them to a broad range of problems. The research concentrates on the design, evaluation, and implementation of cryptographic algorithms and protocols, and on the development of security architectures for information and communication systems. The applications areas are privacy, identity management, anonymous communications, and trusted platforms. Relatively new research topics deal with mass data analysis, forensic multimedia analysis, mobile device security, ransomware, software-defined networks, and 5G. GASS evaluates and develops security for wireless and mobile networks. GASS collaborates with research and development companies (such as Hitachi, USA; IBM Research, USA; INDRA, Spain; Nokia, Spain; Safelayer Secure Communications, Spain; and Treelogic, Spain), institutions (such as the Spanish Ministry of Defense and the Spanish National Police or Science and Engineering Research Support Society, South Korea), and universities (such as the Instituto Politécnico Nacional, Mexico; the Politecnico di Milano, Italy; Saarland University, Germany; the Universidade de Brasília, Brazil; the University of Kent, U.K.; and the University of Tasmania, Australia). His professional experience includes the management of both national and international research projects and both public financing (the Spanish Ministry of Research and Development, the Spanish Ministry of Defence, and the Horizon 2020—European Commission) and private financing (Hitachi, IBM, Nokia, Safelayer Secure Communications, and TB Solutions Security). He has been an advisor of 20 Ph.D. theses in the last ten years. Regarding his scientific activity, he has published numerous publications in prestigious international and conferences. He is also an Editor or a Guest Editor of international journals, such as *Future Generation Computer Systems* (Elsevier), IEEE ACCESS, *Sensors* (MDPI), *Entropy* (MDPI), *Applied Sciences* (MDPI), *The Journal of Supercomputing* (Springer), *IET Communications*, the *International Journal of Ad Hoc and Ubiquitous Computing*, and *Future Internet* (MDPI), among others.